\documentclass[a4paper,12pt,doublespace,authoryear]{article}

\usepackage{graphicx}
\usepackage{amssymb,amsfonts,amsmath}
\usepackage{breakcites}
\usepackage{color}

\usepackage{graphicx}
 \usepackage{mathptmx}   

\def\R{{\mathbb R}}

\usepackage{newtxtext}
\usepackage[varvw]{newtxmath}

\title{A solvable walking model for a two-legged robot}

\author{Rui Dil\~ao and Nuno Teixeira}
\date{}

\begin{document}

\maketitle

\begin{center}
University of Lisbon, Instituto Superior T\'ecnico, Nonlinear Dynamics Group\\
Av. Rovisco Pais, 1049-001 Lisbon, Portugal 
 \end{center}

\begin{center}
ruidilao@tecnico.ulisboa.pt\\
nunowallenstein@gmail.com
\end{center}

\begin{abstract}
We present a solvable biped walking model based on an inverted pendulum with two massless articulated legs capable of walking on uneven floors and inclined planes. The stride of the two-legged robot results from the pendular motion of a standing leg and the articulated motion of a trailing leg. Gaiting is possible due to the alternating role of the legs, the standing and the trailing leg, and the conservation of energy of the pendular motion. The motion on uneven surfaces and inclined planes is possible by imposing the same maximal opening angle between the two legs in the transition between strides and the adaptability of the time of each stride. This model is solvable in closed form and is reversible in time, modelling the different types of biped motion. Several optimisation results for the speed of gaiting as a function of the robot parameters have been derived. 
\end{abstract}


\bigskip

\vfill\eject

\section{Introduction}

Biped human walking and running is a complicated activity involving the coordination of complex motor, sensory and neural systems, \cite{Hol}. The description of all these processes and their coordination is an enormous task demanding detailed knowledge of many biological processes, eventually some of them unknown. The ability of humans to walk requires a long process of learning by trial and error, ultimately associated with the adaptation of several neural mechanisms during human development. 

Although the complexity of human walking, engineers, biologists, physicists, and mathematicians have joined efforts to build machines reproducing human movements and walking. Models describing the basic features of walking and running inspired the industry to create robots to perform some human activities \cite{Ada}. A simple search on the internet leads to different prototype machines mimicking jumping, walking, and running.

Two classes of models were created with entirely different dynamics to reproduce the effects of human walking and running. Walking mechanisms based on an inverted pendulum model were introduced by \cite{Sau, Ale, MM}, while running models based on a spring-mass model were pioneered by \cite{Bli} and \cite{McC}. 
These models have two articulated legs simulating the hip, knees, and feet, with many parameters. These models discussed different mechanisms for transferring kinetic and potential energies between steps. In all the cases, stride dynamics and control are computer-assisted, and all these models are non-solvable in closed form. For a generic comparative review of these models, we refer to \cite{Gey} and \cite{Hol}. 

Here, we introduce a simplified articulated two-legged robotic model with mass concentrated on the hip and feet motion is dynamically absent. This type of model mimics two-legged stilt walking. Some authors refer to this model as a kneed biped robot \cite{Asa} or point foot robot \cite{Che}. For this class of simpler robot models, the different approaches consider knees with mass and elaborated control systems \cite{Asa, Che, Mak}, and stochastically controlled robots \cite{Su}. In these models, stability is critical, and robots should have internal mechanisms for stability control and energy supply to restore the standing leg pendular energy. Moreover, the stride durations in horizontal surfaces are non-constant, and the mathematical models are non-solvable, making the derivation of generic dynamic properties and optimisation goals difficult.
A two-segment robot model based on the inverted pendulum model and able to walk on inclined planes has been introduced by \cite{Nor}. However, due to the absence of articulated knees, this model fails to describe gaiting along uneven surfaces. 
So, we aim to build a solvable model for walking, leading to quantitative predictions about walking dynamics and adaptability to motion on uneven surfaces and inclined planes.

The model introduced here assumes that the energy loss is due to the inelastic contact of the trailing leg with the floor at the end of each stride, and spring-like actuation effects are not considered (\cite{Kuo}). The robotic leg must have an internal supply of energy that restores the pendulum energy, implying that phase space trajectories remain unchanged.
 We further imposed that the angle between the standing and trailing legs at the transition between consecutive strides is fixed, ensuring that gaiting in uneven terrains or inclined planes is possible without complex control mechanisms.

This paper is organised as follows. In section~\ref{sec2}, we build the model for the movement of the two-legged robot along a horizontal flat surface, where the characteristics of the motion and optimisation results as a function of the robot parameter are derived. In section~\ref{sec3}, the model is extended for the motion along uneven surfaces and inclined planes. In the final section, we summarise the conclusions of the paper.

 \section{Motion along a flat surface}\label{sec2}
 
 In the inverted pendulum basic model of locomotion introduced here, the two-legged robot consists of a mass $m$ simulating the hip and two massless articulated legs. One of the legs is the standing leg in contact with the floor, with a pendular movement, and the other articulated trailing leg makes the transition between strides, enabling gaiting. Between consecutive strides, the two legs interchange their roles. To fix parameters, the total length of each leg is $\ell$, $\ell_1$ is the distance from the hip to the knees,  $\ell_2$ is the distance from the knees to the point feet, and $\ell=\ell_1+\ell_2$. In the following two subsections, we describe the motion of each leg separately.

\subsection{The pendular motion of the standing leg}\label{mot1}

\begin{figure}[htbp]
\centering
\includegraphics[width=0.8\textwidth]{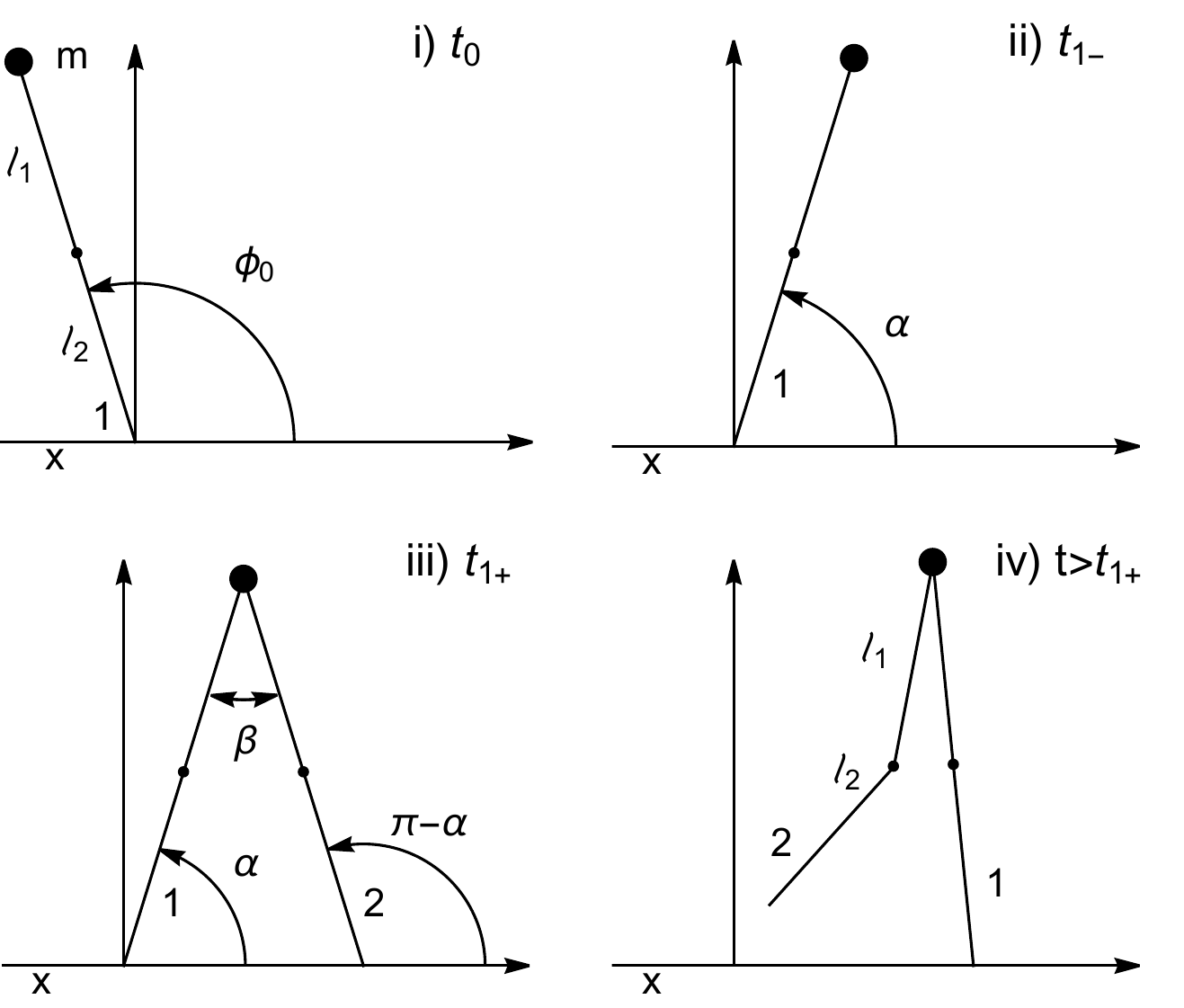}
\caption{Phases of locomotion of a two legged-robot along a horizontal line, in the positive direction of the $x$-axis. (i) Starting position of the standing leg 1 at time $t_0$, making an angle $\phi_0$ with the horizontal direction. (ii) Final position of the pendulum-like motion of the standing leg at time $t_{1-}$. At the end of a stride, the transitions between legs are made with the fixed attack angle $\alpha$, with $0<\alpha <\pi/2$.
 During the first stride, phases i) and ii), the movement of the trailing leg 2 is not represented. iii) Starting time of the second stride at time $t_{1+}$, where the position of the trailing legs 2 is shown. At the time $t_{1+}$, the roles of the two legs are interchanged. iv) The second stride, showing the positions of the two legs at some instant of time $t>t_{1+}$. The articulated motion of the trailing leg 2 is described in subsection~\ref{mot2}.}
\label{figR-1}
\end{figure}

In figure~\ref{figR-1}, the succession of phases of locomotion of the two-legged robot is represented. The different stages of motion have the following characteristics:
\begin{description}
\item{i)} The motion starts at time $t=t_0$, with the standing leg 1 making the angle $\phi(t_0)=\phi_0>\pi/2$ with the horizontal direction, and $0\le \phi (t) \le \pi$. In the first phase of the movement, the initial angular speed of the standing leg 1 is such that $\dot \phi (t_0)<0$, but its absolute value should be large enough for the leg to reach the vertical position at $\phi=\pi/2$. We assume the motion is in the positive direction of the $x$-axis.
\item{ii)} Position of the standing leg at the transition time between strides at $t=t_{1-}>t_0$, where the attack angle $\phi(t_{1-})=\alpha$, with  $0<\alpha <\pi/2$, is a fixed parameter. Between times $t_0$ and $t_{1-}$, the movement of the standing leg is pendular.
\item{iii)} Position of the standing and trailing legs at the instant to time $t=t_{1+}$. The trailing leg upholds on the ground, making an angle $\pi-\alpha$ with the horizontal direction and the standing leg an angle $\alpha$. The trailing leg makes inelastic contact with the ground and dissipates energy. 
At time $t=t_{1+}$, the trailing leg becomes the standing leg ($2\to1$) and \textit{vice versa} ($1\to2$). For the movement to be effective, the two-legged robot must have a source of energy to compensate for the energy dissipated in the inelastic shock with the ground so that the energy of the pendular movement is conserved. The angle $\beta=\pi-2\alpha$ measures the relative maximal opening of the two legs during a stride.
\item{iv)} Position of the standing and trailing legs at some time $t>t_{1+}$.  
The trailing leg has an articulated movement independent of the movement of the standing leg.  During a stride, the leg segment $\ell_1$ of the trailing leg moves in the direction of the motion, and the leg segment $\ell_2$ makes a more complex movement that will be described in subsection~\ref{mot2}.
\end{description}

In polar coordinates, the equation of motion of the standing leg during phases i) and ii) is
\begin{equation}
\ddot \phi+\frac{g}{\ell} \cos \phi=0,
\label{eqR1}
\end{equation}
which has a total energy
\begin{equation}
H=\frac{1}{2}m\ell^2 {\dot \phi}^2+m {g}{\ell} \sin \phi ,
\label{eqR2}
\end{equation}
where $0\le \phi \le\pi$,  $\dot \phi\in \R$  and $g$ is the acceleration of gravity. In the  range of variation of $\phi$, the differential equation (\ref{eqR1}) has only one unstable fixed point with coordinates $(\phi,\dot \phi)=(\pi/2,0)$.

\begin{figure}[h]
\centering
\includegraphics[width=\textwidth]{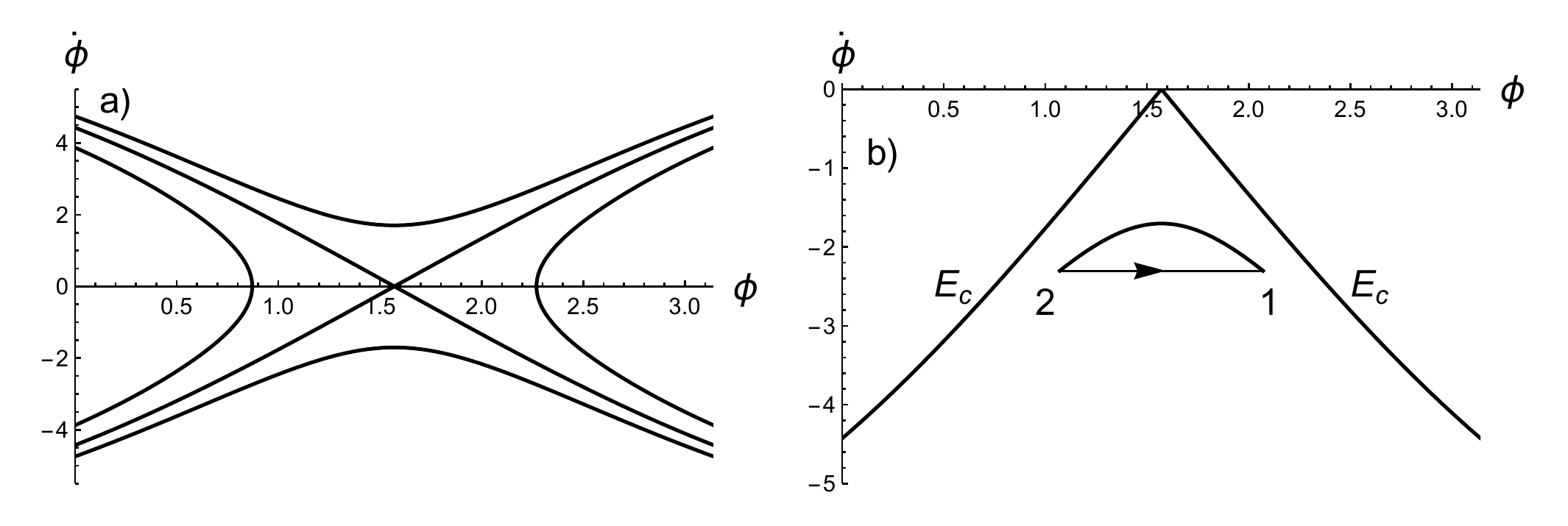}
\caption{a) Phase space curves of equation \eqref{eqR1}, for several energy values $H=E_0$. b) Periodic phase space curve of the standing leg obeying condition \eqref{eqR3}, with $g= 9.8$~ms$^{-2}$, $\ell=1$~m, $m=80$~kg, initial energy $E_0=900$~J, $\alpha=(\pi/2-0.5)$~rad and $\phi_0=\pi/2+0.5=(\pi-\alpha)$~rad, which corresponds to the relative maximal opening of the two legs during a stride $\beta=1$~rad$=57.3^{\circ}$.
The number 1 indicates the initial state of the standing leg, and the number 2 indicates its final state. The arrow indicates the instantaneous change of the trailing to the standing leg, the effect of the energy control mechanism (eq. \eqref{eqR6a}-\eqref{eqR6b}), and the direction of circulation in the phase space. The phase space curves with energy $E_c=mg\ell=784$~J, defining the limiting energy necessary for the existence of gaiting, are also represented.}
\label{figR-2}
\end{figure}

As the inverted pendulum equation \eqref{eqR1} is Hamiltonian, the energy is conserved, and the phase space curves have the equation $\dot \phi=\pm \sqrt{2 E_0/(m \ell^2)-2 g \sin \phi/\ell}$, where $E_0$ is the initial energy of the two-legged robot. If $E_0\ge mg\ell:=E_c$, the phase space curves are defined for $\phi\in[0,\pi]$. If $0\le E_0< E_c$, the phase space curves are defined for $\phi\in [0,\arcsin(E_0/E_c]\cup [\pi-\arcsin( E_0/E_c,\pi]$. In figure~\ref{figR-2}a),  we show the phase space curves of the equation (\ref{eqR1}) of the inverted pendulum.

For the two-legged robot to start to move and reach the vertical position, the energy deposited in the initial conditions of the pendulum has to obey the condition $E_0> E_c$. To complete phases i)-iii) in figure~\ref{figR-1}, by  (\ref{eqR2}), this energy condition leads to
\begin{equation}
{\dot \phi_0}^2>2\frac{g}{\ell} (1-\sin \phi_0 ).
\label{eqR3}
\end{equation}
In the case of planar motion and
for simplicity, it can be assumed that $\phi_0=(\pi-\alpha)$, and therefore the condition \eqref{eqR3} reduces to ${\dot \phi_0}^2>2\frac {g}{\ell} (1-\cos \beta/2 )$, where $\beta=\pi-2\alpha$.

We now calculate the angular speed of the trailing leg at the instant of time 
  $t_{1+}$. By \eqref{eqR2}, with $E_0$ being the initial energy of the standing leg, by conservation of energy, the angular velocity of the trailing leg is  
\begin{equation}
{\dot \phi}(t_1+)={\dot \phi}(t_1-)=-\sqrt{2E_0/(m\ell^2)-2\frac{g}{\ell} \sin \alpha},
\label{eqR4}
\end{equation}
where 
the angle ${\phi}(t_{1-})$ is measured relative to the contact point of the standing leg, and ${\phi}(t_{1+})$ is measured relative to the contact point of the trailing leg.

If the initial conditions obey the inequality (\ref{eqR3}), due to the relationship \eqref{eqR4}, the robot has a periodic movement along the positive direction of the $x$-axis (figure~\ref{figR-2}b).

By (\ref{eqR2}), the duration of one stride  is
\begin{equation}
\begin{array}{lcl}\displaystyle
T(\phi_0,\alpha,E_0)&=&\displaystyle -\int_{\phi_0}^{\alpha}\frac{d\phi}{\sqrt{2E_0/(m\ell^2)-2\frac{g}{\ell} \sin \phi}}\\ \displaystyle
&=&\displaystyle
\frac{2}{A}\left(F\left(\frac{\pi-2\alpha}{4} ,-4\frac{g/\ell}{A^2}\right)-F\left(\frac{\pi-2\phi_0}{4},-4\frac{g/\ell}{A^2}\right)\right),\end{array}
\label{eqR5b}
\end{equation}
where $A=\sqrt{2E_0/(m\ell^2)-2g/\ell}$ and $F(x,m)$ is the elliptic integral of the first kind. For the case $\phi_0=(\pi-\alpha)$,
\begin{equation}
T(\pi-\alpha,\alpha,E_0)=\frac{4}{A}F\left(\frac{\pi-2\alpha}{4},-4\frac{g/\ell} {A^2}\right).
\label{eqR5}
\end{equation}
As each stride has length $a_0=2 \ell \cos \alpha $,  the two-legged robot speed during a stride is
\begin{equation}
v_m=a_0/T(\pi-\alpha,\alpha ,E_0)=\frac{\ell \cos \alpha}{2}A\ F\left(\frac{\pi-2\alpha}{4},-4\frac{g/\ell} {A^2}\right).
\label{eqR5c}
\end{equation}

\begin{figure}[h]
\centering
\includegraphics[width=\textwidth]{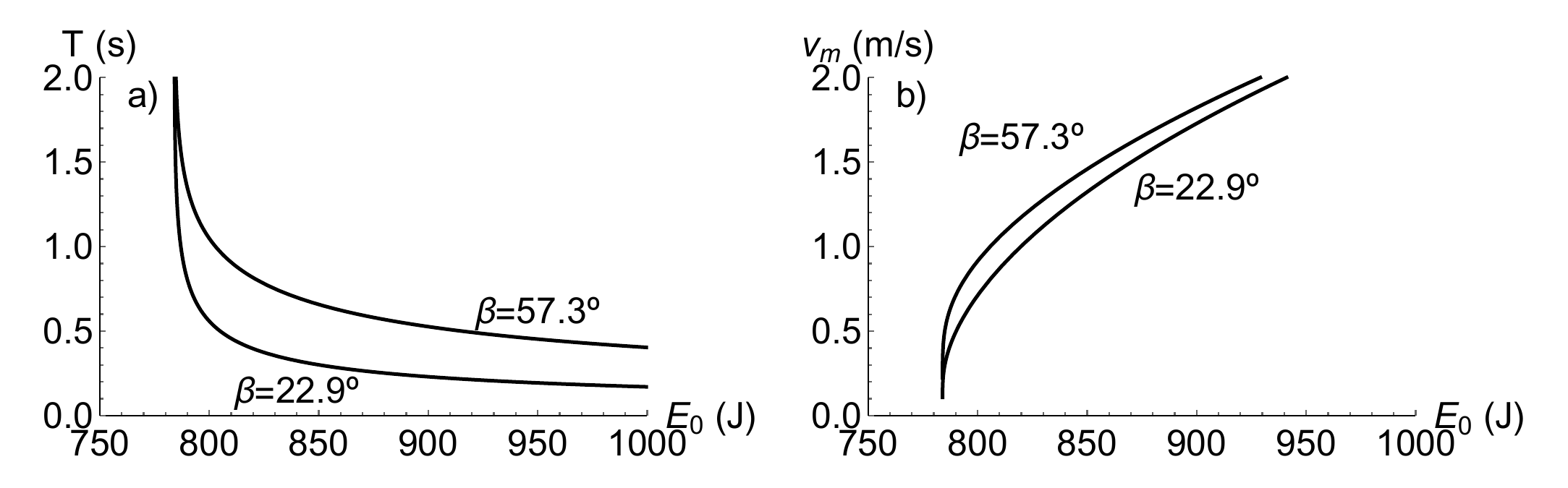}
\caption{a) Duration of a stride of the two-legged robot model along a horizontal surface as a function of energy $E_0$, calculated from \eqref{eqR5},  for several values of $\beta$.  b) Speed of a stride as a function of energy $E_0$, calculated from \eqref{eqR5c}, for several values of $\beta$. The simulation parameters are $\phi_0=(\pi-\alpha)=(\pi+\beta)/2$, $g= 9.8$~ms$^{-2}$, $ m=80$~kg and $\ell=1$~m.}
\label{figR-3}
\end{figure}

In figure~\ref{figR-3}, we show the duration of the stride of the two-legged robot model along a horizontal surface, as a function of the energy $E_0$ and its speed, for several values of the maximum opening of the legs $\beta $. This shows that the speed of a stride increases as the energy $E_0$ or the maximal opening angle $\beta$ increases.

\begin{figure}[h]
\centering
\includegraphics[width=0.6\textwidth]{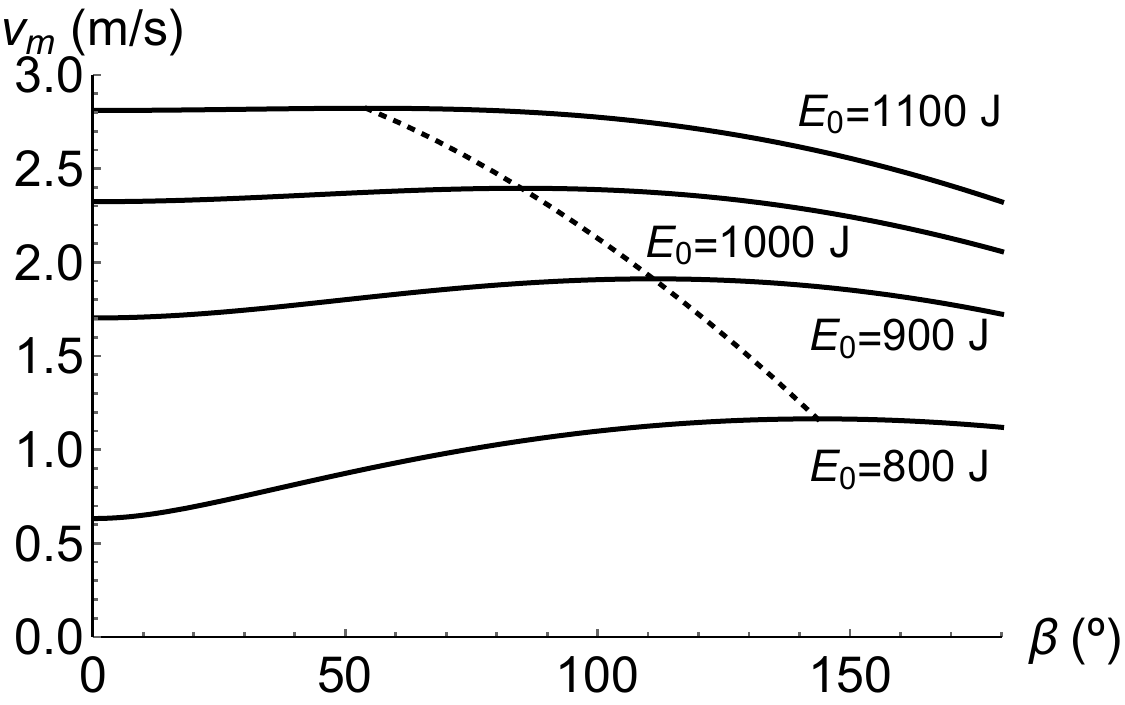}
\caption{Speed of a two-legged robot as a function of the opening of the two legs $\beta $, for several values of the initial energy $E_0$. The dashed line indicates the maximum value for the speed of the two-legged robot. The simulation parameters are  $\phi_0=(\pi-\alpha)=(\pi+\beta)/2$, $g= 9.8$~ms$^{-2}$, $ m=80$~kg and $\ell=1$~m.}
\label{figR-4}
\end{figure}

In figure~\ref{figR-4}, we show the speed of the two-legged robot as a function of the opening of the legs $\beta$, for several values of the initial energy $E_0$. So, for each initial energy $E_0$, there is an angular opening of the legs that maximises the speed of the two-legged robot. The greater the energy $E_0$, the smaller the angular opening of the legs to maximise the two-legged robot speed.

For gaiting to be effective, we must impose conservation of energy in the transition of the trailing leg to the supporting leg ($t_{1-}\to t_{1+}$). The two-legged robot must have an energy supply system to compensate for the energy lost during the impact with the floor of the trailing leg. This impact is assumed to be inelastic. 

At the transition from the trailing to the standing leg, the velocity of the hip mass $m$ is
\begin{equation}
\begin{array}{lcl}\displaystyle
{\bf v}(t_{1-})&\displaystyle=&\displaystyle \ell \dot \phi(t_{1-}) (-\sin \phi(t_{1-})\   {\bf e}_x+\cos \phi(t_{1-})\ {\bf e}_y)\\ \displaystyle
&\displaystyle =&\displaystyle \ell \dot \phi(t_{1-}) (-\sin \alpha\ {\bf e}_x+\cos \alpha\ {\bf e}_y).
\end{array}
\label{vel1}
\end{equation}
In the collision with the floor, energy is lost, and the component of ${\bf v}(t_{1-})$ along the direction of the trailing leg $2$ is absorbed by the ground (inelastic shock). At the instant of time $t_{1+}$, the direction of the trailing leg $2$ is
$$
{\bf d}=\cos(\pi-\alpha)\ {\bf e}_x+\sin(\pi-\alpha)\ {\bf e}_y.
$$
As the projection of ${\bf v}(t_{1-})$ along ${\bf d}$ is ${\bf v}(t_{1-}).{\bf d}$, immediately after the shock of the trailing leg with the floor, the velocity of the mass $m$ of the hip is
\begin{equation}
\begin{array}{lcl}
{\bf v}'&=&{\bf v}(t_{1-})-({\bf v}(t_{1-}).{\bf d})\ {\bf d}\\
&=& \ell \dot \phi(t_{1-}) \left(\sin \alpha \cos 2\alpha \ {\bf e}_x+\cos \alpha \cos 2\alpha\ {\bf e}_x\right).
\end{array}
\label{vel2}
\end{equation}
So, and as $\dot \phi(t_{1-}) <0$, we have two cases
$$
\begin{array}{lcl}
v_x'\ge 0, \ v_y'\ge 0 &\quad \hbox{for}\quad & \pi/4\le \alpha <\pi/2\quad \hbox{or}\quad \beta \le\pi/2\\
v_x'<0,\ v_y'<0 &\quad \hbox{for}\quad & 0\le \alpha <\pi/2\quad \hbox{or}\quad \beta >\pi/2.
\end{array}
$$
In the first case, the hip velocity is in the direction of motion after the shock. Still, in the second case, the velocity is directed opposite to the motion. Thus, for the energy of the pendular movement to be conserved in the transitions between strides, it is necessary to compensate for the energy lost in the shock by some mechanism internal to the two-legged robot.
Let $E_p$ be the energy to be replenished in the transition between strides.

If $(v_x'\ge 0, v_y'\ge 0)$, by \eqref{vel1}, \eqref{vel2} and \eqref{eqR4}, then
\begin{equation}
\begin{array}{lcl}\displaystyle
E_p&=& \frac{1}{2}mv^2-\frac{1}{2}m \vec v'^2= \frac{1}{2}m\ell^2 \dot \phi(t_1 -)^2 (1-\cos^2 (2 \alpha))\\\displaystyle
\displaystyle &=&\displaystyle (E_0-mg \ell \sin \alpha )\sin^2 (2 \alpha)=(E_0-mg \ell \cos \beta/2 )\sin^2 ( \beta).
\end{array}
\label{eqR6a}
\end{equation}

If $(v_x'<0,\ v_y'<0)$, as the velocity after the impact has a direction contrary to the direction of the movement and, to maintain gaiting, it is necessary to supply to the mass $m$ the energy
\begin{equation}
\begin{array}{lcl}\displaystyle
E_p&=&E_0-mg\ell \sin\alpha+\frac{1}{2}m v'^2=E_0-mg\ell \sin\alpha+ \frac{1}{2}m\ell^2 \dot \phi(t_1-)^2 \cos^2 (2 \alpha)\\
\displaystyle &=&\displaystyle E_0-mg\ell \sin\alpha+(E_0-mg \ell \sin \alpha )\cos^2 (2 \alpha)\\
\displaystyle &=&\displaystyle (E_0-mg \ell \cos \beta/2 )(1+\cos^2 (\beta)).
\end{array}
\label{eqR6b}
\end{equation}

\begin{figure}[h]
\centering
\includegraphics[width=0.6\textwidth]{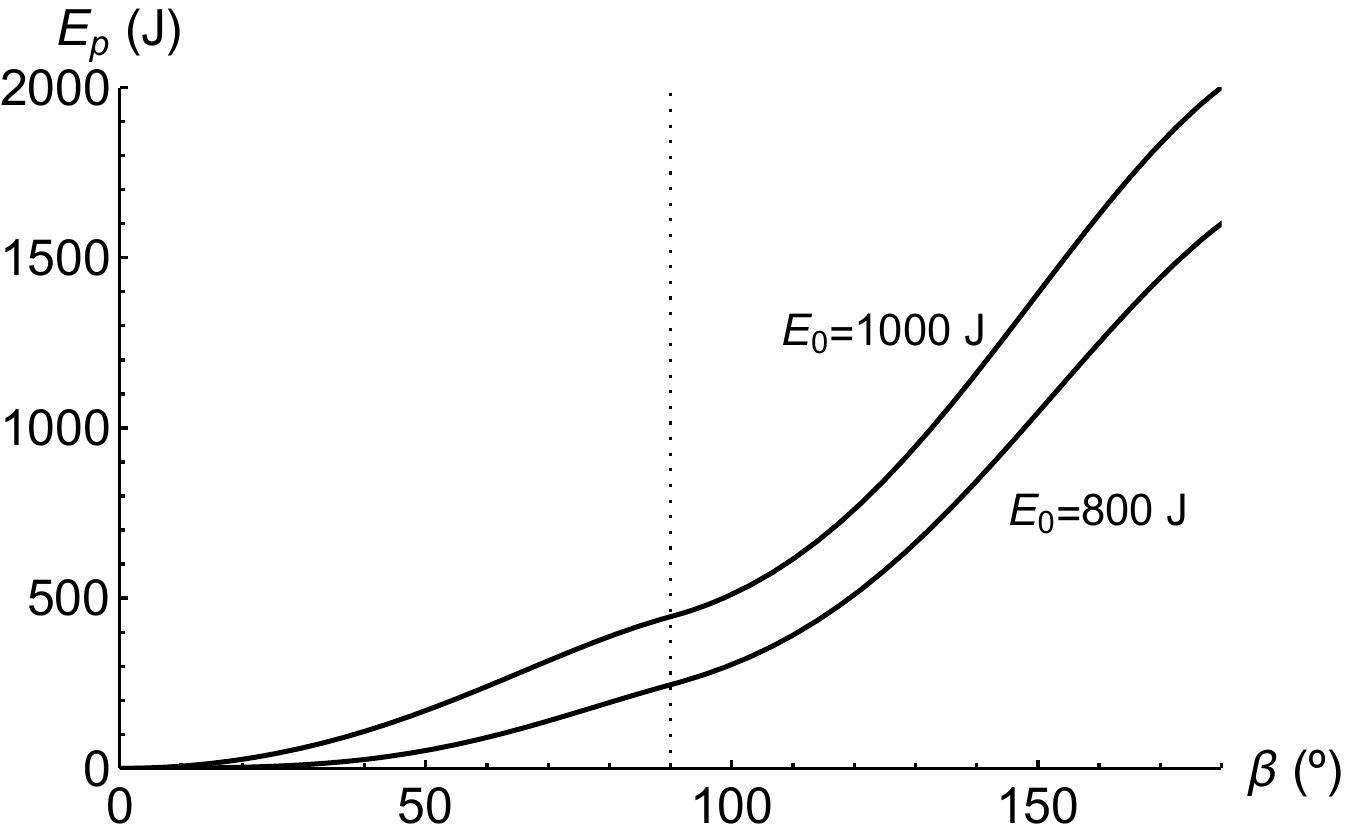}
\caption{Energy necessary to maintain a stride on a flat surface at the transition   $t_{1-}\to t_{1+}$, as a function of $\beta=\pi-2\alpha$. $E_p$ has been calculated by \eqref{eqR6a} and \eqref{eqR6b}.
The simulation parameters are $\phi_0=(\pi-\alpha)=(\pi+\beta)/2$, $g= 9.8$~ms$^{-2}$, $ m=80$~kg and $\ell=1$~m.}
\label{figR-5}
\end{figure}

In figure~\ref{figR-5}, we show the energy $E_p$ necessary to maintain a stride on a flat surface at the transition time $t_{1-}\to t_{1+}$, as a function of $\beta$. Therefore, the larger the stride, the larger the energy replenishment. 

Due to the structure of phase space orbits shown in figure~\ref{figR-2}b) and assuming that the energy supply is replenished at each stride,  the two-legged robot model presented here is stable, as slight variations in the initial conditions and robot parameters lead to neighbouring closed trajectories in phase space.

\subsection{Trailing leg movement}\label{mot2}

In this model of a robotic leg, during phases i)-iii) of figure~\ref{figR-1}, 
the trailing leg has a movement that cannot collide with the floor, and its function alternates with the standing leg.

\begin{figure}[ht]
\centering
 \includegraphics[width= 0.8\hsize]{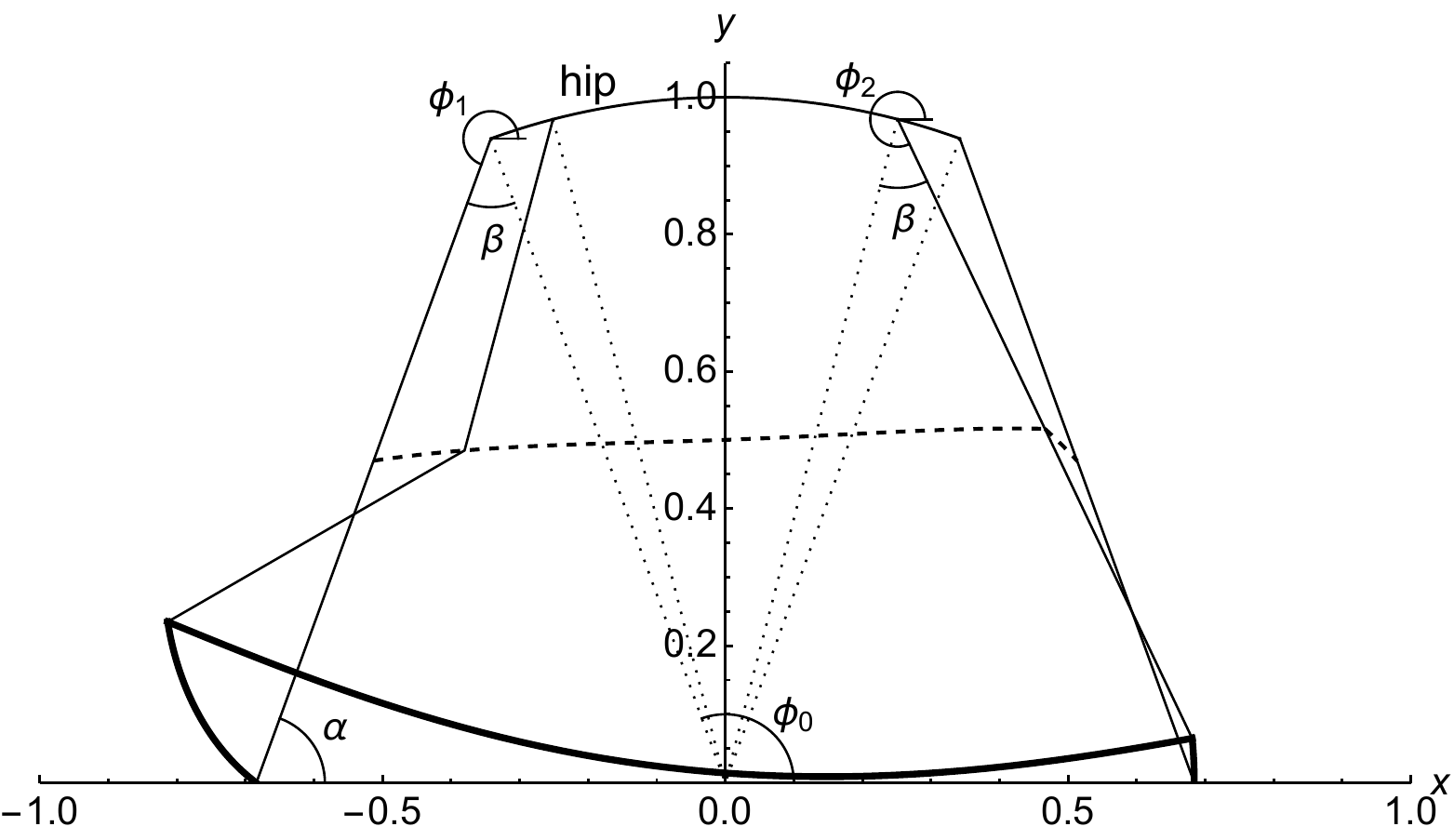}
 \caption{Motion of a two-legged robot during one step. Thin lines represent the movement of the trailing leg. Dotted lines represent the movement of the standing leg. The three phases of the movement of the foot tip of the trailing leg are shown by thick lines. The dashed lines show the knee movement. The retraction and progression movements of the trailing leg foot tip are necessary to maintain a periodic planar motion so that the foot tip does not have contact with the floor. On flat surfaces,  the duration of a stride is $T(\phi_0,\alpha, E_0)$. The first phase of the foot movement of the trailing leg has duration $\gamma_f T$, the second phase has duration $(\gamma_k-\gamma_f) T$ and the third phase has duration $(1-\gamma_k) T$. The simulation parameters are $(x_0,y_0)=(0,0)$, $\phi_0=110^{\circ}$, $\alpha=70^{\circ}$, $E_0=800$~J, $g= 9.8$~ms$^{-2}$, $m=80$~kg, $\ell=1$~m, $\ell_1=0.5$~m, $ \gamma_f=0.1$, $\gamma_k=0.8$ and $\alpha_r=-40^{\circ}$.}
\label{figR-6}
\end{figure}

We further impose that, during one stride, the movement of the foot tip of the trailing leg has three phases, as shown in figure~\ref{figR-6}. In the first phase, the foot of the trailing leg retracts in the opposite direction of motion, while the knee and hip progress in the direction of movement. In the second phase, the foot of the trailing leg moves in the direction of motion until the trailing leg is wholly stretched and makes an angle $\beta$ relative to the standing leg. Finally, the foot of the trailing leg rotates backwards until the foot tip touches the ground. During this movement, the knee of the trailing leg always moves in the direction of motion. 
During one stride, the knee of the trailing leg has two phases. 

To describe the movement of the trailing leg during a stride, we consider the rotation matrix 
\begin{equation}
R(\omega ,t)=\begin{pmatrix}
\cos \omega t & -\sin \omega t \\
\sin \omega t & \cos \omega t
\end{pmatrix}.
\label{eq2.5}
\end{equation}
Let $t_0$ (figure~\ref{figR-1}) be the initial time of a stride and $(x_0,y_0)$ the coordinates of the support point of the standing leg. The positions of the hip, knee and trailing leg foot tip are   $(x_h,y_h)$, $(x_k,y_k)$ and $(x_f ,y_f)$, respectively.

{\bf First phase of the movement of the knee of the trailing leg.}  
During a stride, the first phase of the movement of the knee of the trailing leg has coordinates
\begin{equation}
\begin{pmatrix}{x}_k(t)\\ {y}_k(t)\end{pmatrix}=
\begin{pmatrix}{x}_h(t)\\ {y}_h(t)\end{pmatrix}+\ell_1 R(\omega_k,t-t_0)
\begin{pmatrix} \cos \phi_1\\ \sin \phi_1\end{pmatrix},
\label{eq2.6}
\end{equation}
where ${x}_h(t)=x_0+\ell \cos\phi(t)$, ${y}_h(t)=y_0+\ell \sin\phi(t)$ and $\phi(t) $ is solution of equation \eqref{eqR1}, for the initial condition $\phi_0=\phi(t_0)$ and $(t-t_0)\in [0 ,\gamma_k T(\phi_0,\alpha, E_0)]$.
The angle $\phi_1=(\pi+\alpha)$ is the orientation of the trailing leg at the beginning of the stride, measured relative to the position of the hip (figure~\ref{figR-6}).
The new parameter $\gamma_k <1$ describes the duration of the first movement of the knee of the trailing leg, measured as a fraction of the time of one step. Consider the new angle
\begin{equation}
\phi_2=\pi+\beta+\arctan ((y_h(\gamma_k T(\phi_0,\alpha, E_0))-y_0)/(x_h(\gamma_k T(\phi_0,\alpha, E_0))-x_0),
\label{eq2.6b1}
\end{equation}
describing, at time $t=\gamma_k T(\phi_0,\alpha, E_0)$, the orientation of the segment $\ell_1$ of the trailing leg, measured relative to the position of the hip.
Thus, from the condition $\omega_k\gamma_k T(\phi_0,\alpha, E_0)=(\phi_2-\phi_1)$ (figure~\ref{figR-6}), it follows that
\begin{equation}
\omega_k=(\phi_2-\phi_1) /(\gamma_k T(\phi_0,\alpha, E_0))>0,
\label{eq2.6b}
\end{equation}
where $\omega_k$ and $\phi_2$ are determined by the choices of $\gamma_k$ and the remaining stride parameters. In figure~\ref{figR-6}, in dashed lines, we represent the first phase of movement of the knee of the trailing leg,  calculated by \eqref{eq2.6}-\eqref{eq2.6b}.

{\bf First and second phase of the movement of the trailing leg foot tip.}
During the first phase of the movement of the foot of the trailing leg, its coordinates are
\begin{equation}
\begin{pmatrix}{x}_{f1}(t)\\ {y}_{f1}(t)\end{pmatrix}=
\begin{pmatrix}{x}_k(t)\\ {y}_k(t)\end{pmatrix}+\ell_2 R(\omega_{f1},t-t_0)
\begin{pmatrix} \cos (2\pi-\phi_0)\\ \sin (2\pi-\phi_0)\end{pmatrix},
\label{eq2.6c}
\end{equation}
where $\omega_{f1}<0$ is a parameter to be determined, the new parameter $\gamma_f$ is such that $0<\gamma_f<\gamma_k$, and $(t-t_0)\in [0,\gamma_f T(\phi_0,\alpha, E_0)]$.
The maximum angle of retraction of the foot of the trailing led, measured relative to the initial position of the standing leg, is
\begin{equation}
-\pi<\alpha_{r}<0.
\label{eq2.6d}
\end{equation}
Thus, from the condition $\omega_{f1}\gamma_f T(\phi_0,\alpha, E_0)=\alpha_r$, it follows that
\begin{equation}
\omega_{f1}= \alpha_r /(\gamma_f T(\phi_0,\alpha, E_0))<0,
\label{eq2.6e}
\end{equation}
where $\omega_{f1}$ is determined by the choices of $\gamma_f$, $\alpha_r$ and the remaining stride parameters (figure~\ref{figR-6}).

 In the time interval $t-t_0\in [\gamma_f T(\phi_0,\alpha, E_0),\gamma_kT(\phi_0,\alpha, E_0)]$ --- second phase of the movement of the foot of the trailing leg, the foot of the trailing leg has the trajectory
\begin{equation}
\begin{pmatrix}{x}_{f2}(t)\\ {y}_{f2}(t)\end{pmatrix}=
\begin{pmatrix}{x}_k(t)\\ {y}_k(t)\end{pmatrix}+\ell_2 R(\omega_{f2},t-\gamma_f T(\phi_0,\alpha, E_0 ))
\begin{pmatrix} \cos (\phi_1+\alpha_r)\\ \sin (\phi_1+\alpha_r)\end{pmatrix},
\label{eq2.6f}
\end{equation}
where
\begin{equation}
\omega_{f2}=(\phi_2-\phi_1-\alpha_r) /((\gamma_k-\gamma_f) T(\phi_0,\alpha, E_0))>0.
\label{eq2.6g}
\end{equation}

In figure~\ref{figR-6}, the thick line represents the foot movement of the trailing leg foot tip, calculated through \eqref{eq2.6c}-\eqref{eq2.6g}.

{\bf Second phase of movement of the knee and third phase of movement of the foot of the trailing leg.}
In this case, the movement occurs during the time interval $t-t_0\in [\gamma_k T(\phi_0,\alpha, E_0),T(\phi_0,\alpha,E_0)]$ and the knee  has coordinates
\begin{equation}
\begin{pmatrix}{x}_{k2}(t)\\ {y}_{k2}(t)\end{pmatrix}=
\begin{pmatrix}{x}_h(t)\\ {y}_h(t)\end{pmatrix}+\ell_1
\begin{pmatrix} \cos (\phi_3(t))\\ \sin (\phi_3(t))\end{pmatrix},
\label{eq2.7}
\end{equation}
where
$$
\phi_3(t)=\pi+\beta+\arctan((y_h(t)-y_0)/(x_h(t)-x_0)).
$$
Likewise, the movement of the foot has coordinates
\begin{equation}
\begin{pmatrix}{x}_{f3}(t)\\ {y}_{f3}(t)\end{pmatrix}=
\begin{pmatrix}{x}_h(t)\\ {y}_h(t)\end{pmatrix}+\ell
\begin{pmatrix} \cos (\phi_3(t))\\ \sin (\phi_3(t))\end{pmatrix}.
\label{eq2.7b}
\end{equation}

\begin{figure}
\centering
 \includegraphics[width= 0.49\hsize]{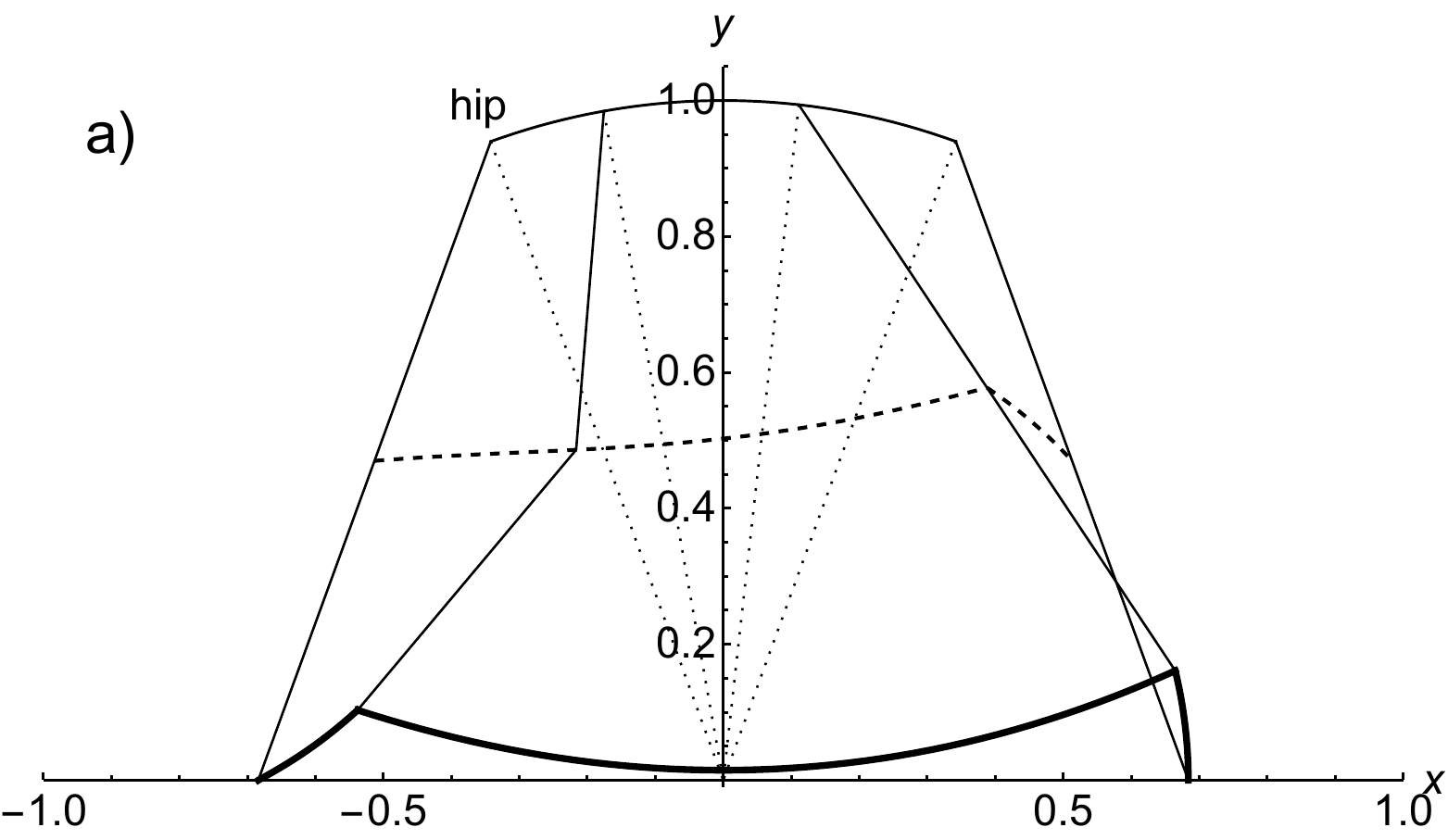} 
 \includegraphics[width= 0.49\hsize]{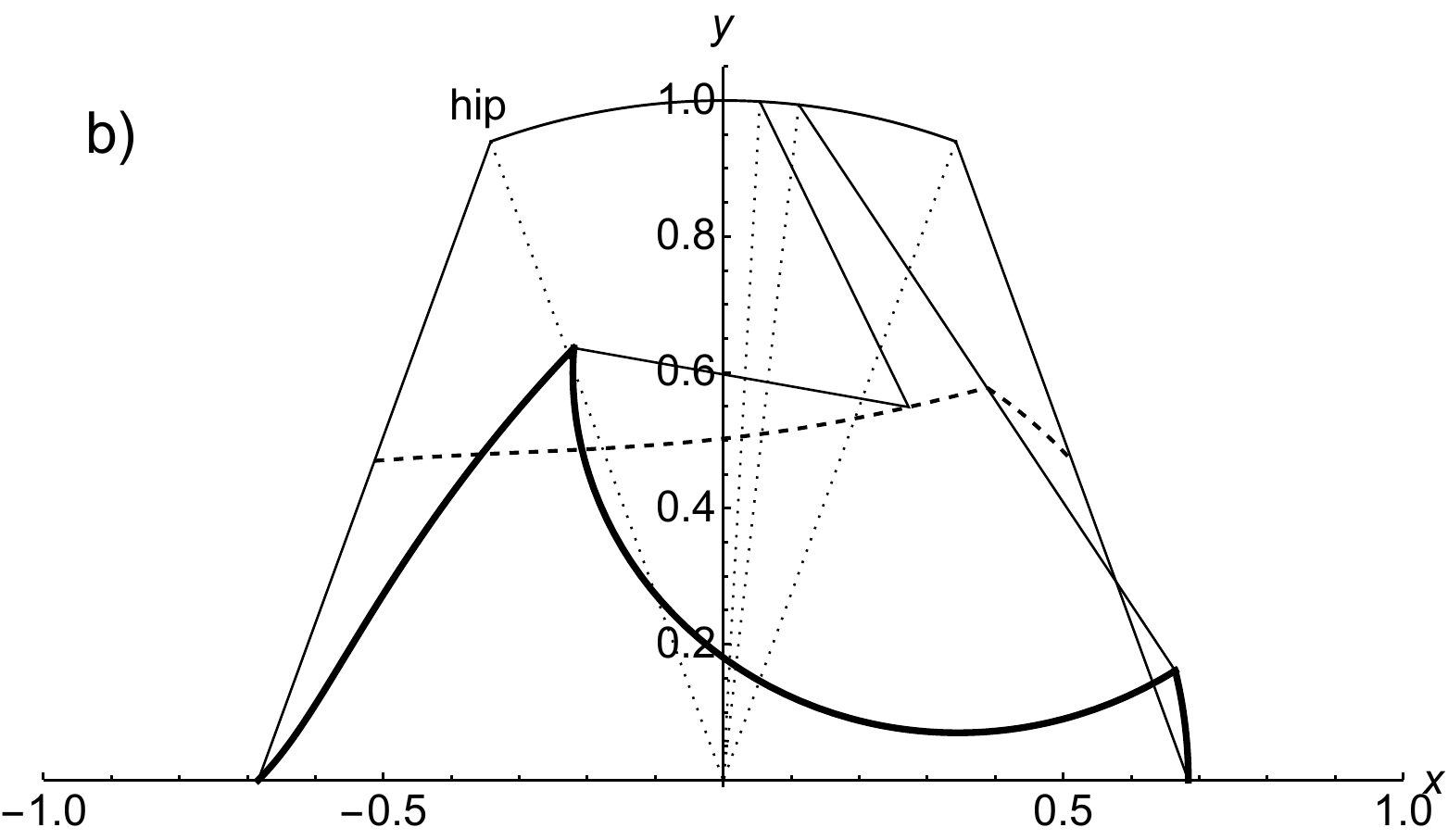}
 \caption{Motion of the foot tip of the trailing leg showing no retraction. The simulation parameters are $(x_0,y_0)=(0,0)$, $\phi_0=110^{\circ}$, $\alpha=70^{\circ}$, $E_0=800$~J, $g= 9.8$~ms$^{-2}$, $m=80$~kg, $\ell=1$~m, $\ell_1=0.5$~m; a) $\gamma_f=0.2$, $\gamma_k=0.7$ and $\alpha_r=-20^{\circ}$; b) $\gamma_f=0.6$, $\gamma_k=0.7$ and $\alpha_r=-80^{\circ}$. If $\gamma_f=\gamma_k<1$, the second phase of the movement of the foot of the trailing leg is instantaneous.}
\label{figR-7}
\end{figure}

A sufficient condition for the effectiveness of a stride of the two-legged robot is that  $y_{f1}(t)>0$, for every $t\in (t_0,\gamma_fT(\phi_0,\alpha, E_0)]$, and $y_{f2}(t)\ge 0$, for every
$t\in [\gamma_f T(\phi_0,\alpha, E_0),\gamma_k T(\phi_0,\alpha, E_0)]$. 
These two conditions depend on the choice of the stride parameters. There can be no retraction in the first phase of the foot tip of the trailing leg. By making a different choice of parameters, the foot movement of the trailing leg does not necessarily have a retraction movement (figure~\ref{figR-7}), although $\alpha_r<0$.
 
\begin{figure}
\centering
 \includegraphics[width= 0.6\hsize]{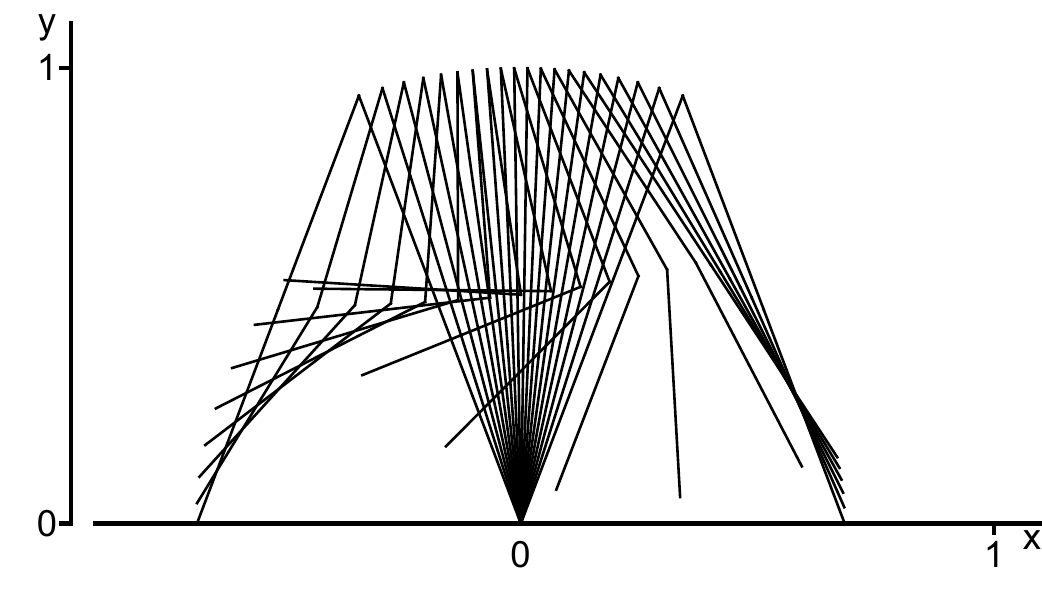}
 \caption{Temporal evolution of the position of the two-legged robot during one stride. As the surface is flat, this stride extends to a gaiting motion. The simulation parameters are $(x_0,y_0)=(0,0)$, $\phi_0=110^{\circ}$, $\alpha=70^{\circ}$, $E_0=800$~J, $g= 9.8$~ms$^{-2}$, $m=80$~kg, $\ell=1$~m, $\ell_1=0.5$~m, $ \gamma_f=0.4$, $\gamma_k=0.7$ and $\alpha_r=-80^{\circ}$. The duration of a stride is $T=0.84$~s, and its length is $0.68$~m.}
\label{figR-8}
\end{figure}

In any of the analysed cases, the two-legged robot motion on a flat surface is periodic and valid for all $t\in \R$. To visualise the movement of the two-legged robot, in figure~\ref{figR-8}, we show the positions of the two legs during one stride on a flat surface.

In the Supplementary Material, we present a video of the motion of the two-legged robot on a horizontal surface (robot-flat.mp4), showing the reversibility of the motion. 

\begin{figure}
\centering
 \includegraphics[width= \hsize]{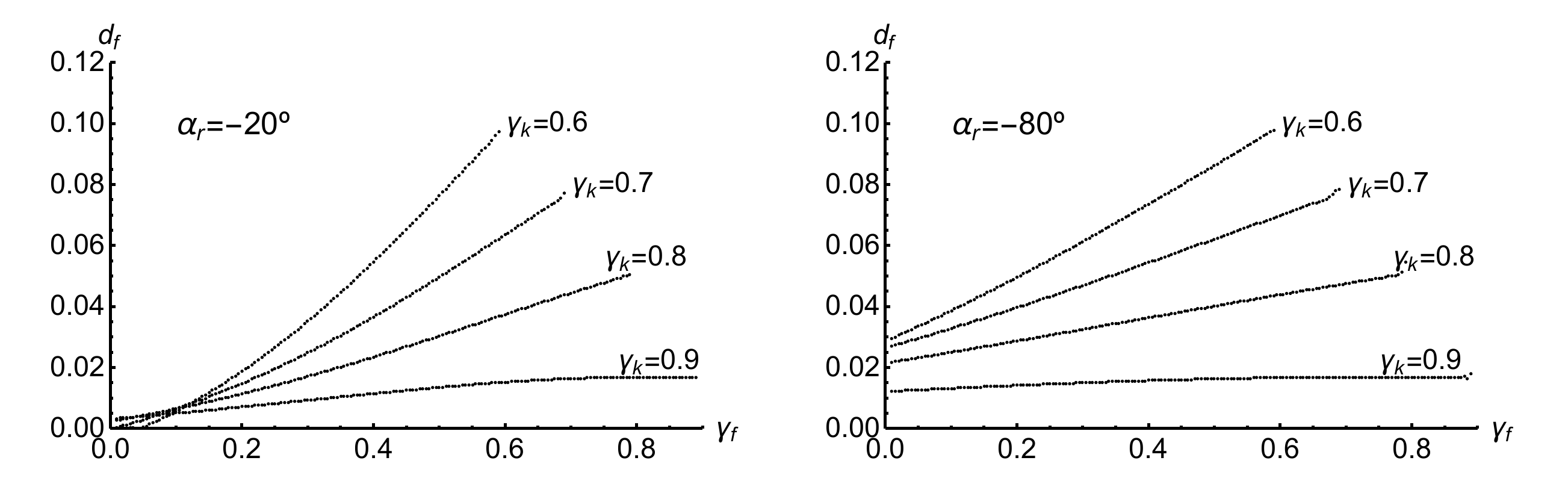}
 \caption{Minimum distance from the foot tip of the trailing leg to the floor \eqref{eq-df}, during one stride, as a function of $\gamma_f$, for several values of $\gamma_k$ and $\alpha_r$. The remaining parameters of the simulation are $(x_0,y_0)=(0,0)$, $\phi_0=110^{\circ}$, $\alpha=70^{\circ}$, $E_0=800$~J, $g= 9.8$~ms$^{-2}$, $m=80$~kg, $\ell=1$~m and $\ell_1=0.5$~m.}
\label{figR-9}
\end{figure}

The stride is expected to adapt to the motion on uneven surfaces with the model just developed for horizontal surfaces. Thus, the minimum distance from the foot of the trailing leg to the floor
\begin{equation}
d_f(\gamma_f,\gamma_k)=\min_{t\in [\gamma_f T, \gamma_k T]}y_{f2}(t)
\label{eq-df}
\end{equation}
has been analysed as a function of $\gamma_f$ and for various values of $\gamma_k$ and $\alpha_r$. 
 As shown in figure~\ref{figR-9},  in some cases, there is a choice of the parameters $\gamma_f$, $\gamma_k$ and $\alpha_r$ to keep the foot tip of the trailing leg away from the floor.

\section{Motion on non-planar surfaces and inclined planes}\label{sec3}

To reach the final support position without touching the floor, the trailing leg is articulated at the knee  (figures~\ref{figR-7} and \ref{figR-8}). The same model can be applied to consecutive strides on non-planar surfaces, described by continuous and differentiable functions $y= f(x)$, with $x\in \R$,  deviating slightly from the horizontal line. Let 
\begin{equation}
d_{sr}=\max_{x\in \R} f(x)-\min_{x\in \R} f(x)<\infty
\label{rough}
\end{equation}
be the surface roughness parameter. Assume that $d_{sr}$ is less than the minimum distance from the foot of the trailing leg to the ground during the second phase of the movement of the foot of the trailing leg. That is,
\begin{equation}
d_{sr}<d_f=\min_{t\in [\gamma_f T, \gamma_k T]}y_{f2}(t)
\label{rough2}
\end{equation}

Imposing further that $|f'(x)|<\tan \alpha$, for all $x\in \R$, the standing leg can not have another contact with the floor except at the supporting point. 

 \begin{figure}[htbp]
\centering
\includegraphics[width=0.5\textwidth]{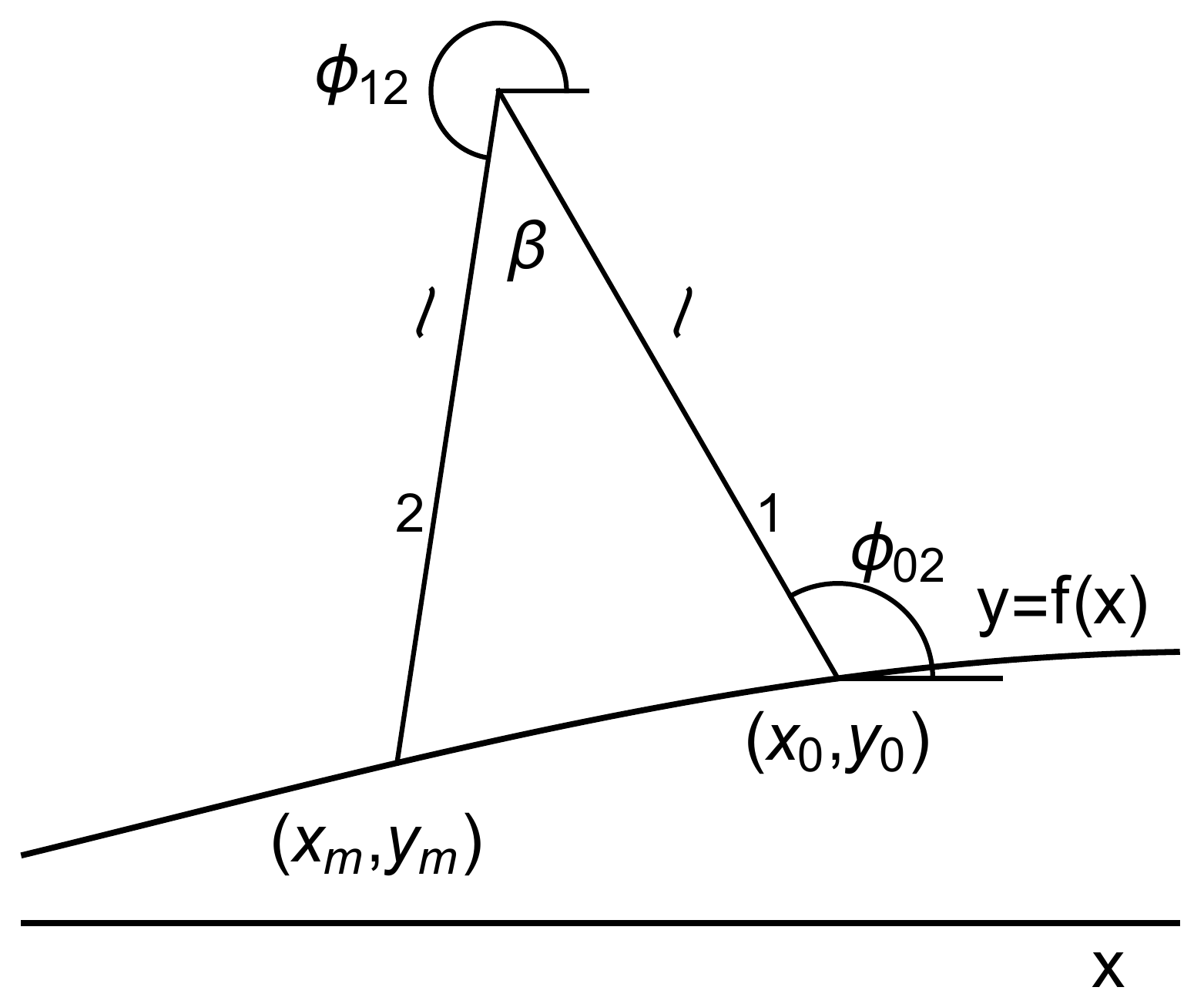}
\caption{Position of the two-legged robot on a non-horizontal surface described by the function $y=f(x)\ge0$. When the two feet are simultaneously supported, the robotic leg forms an isosceles triangle with angles $\alpha$ and $\beta$, where $\beta+2\alpha=\pi$. The distance between the two feet are $\ell_b=\ell\sqrt{2-2\cos\beta}$.}
\label{figR-10}
\end{figure}
 
 Consider that the movement on a non-planar surface is described by a continuous and differentiable function $y=f(x) \ge 0$, with $|f'(x)|<\tan \alpha$. In the transition between strides, the two legs of the robot are supported on a non-planar floor, and the robot has the geometry of an isosceles triangle with angles such that $\beta+2\alpha=\pi$ (figure~\ref{figR-10}).
 Let $(x_0,y_0=f(x_0))$ be the initial coordinates of the foot of the standing leg. By a simple calculation, the side of the triangle opposite to the angle $\beta$ has length $\ell_b=\ell\sqrt{2-2\cos\beta}$. Then, the initial coordinates of the foot of the trailing leg are
\begin{equation}
\left\{
\begin{array}{l}
x_m=x_0+\ell_b\cos (\phi_{02}+\alpha)\\
y_m=y_0+\ell_b\sin (\phi_{02}+\alpha),
\end{array}\right.
\label{eq-sup1}
\end{equation}
where $\phi_{02}$ is an angle dependent on the shape of the surface (figure~\ref{figR-10}), determined by the condition
\begin{equation}
y_0+\ell_b\sin (\phi_{02}+\alpha)=f(x_0+\ell_b\cos (\phi_{02}+\alpha)).
\label{eq-sup2}
\end{equation}
The angle $\phi_{02}$, determined for non-planar surfaces, replaces the angle $\phi_0$ of the figure~\ref{figR-1} (figure~\ref{figR-10}). Therefore, the initial angle of the trailing leg, measured relative to the hip position, is
\begin{equation}
\phi_{12}=\pi+\phi_{02}-\beta . 
\label{eq-sup3}
\end{equation}

The motion of the robotic leg starts with the position of both feet determined by \eqref{eq-sup1} -- \eqref{eq-sup3}. At the beginning of each step, the angle $\phi_{12}$ replaces the angle $\phi_{1}$ in \eqref{eq2.6}. By \eqref{eqR3}, for the movement to be feasible, we must have ${\dot \phi_{02}}^2>2 g  (1- \sin \phi_{02} )/\ell$.

Starting from the initial positions of the two legs with fixed $\beta$, the movement proceeds according to the rules determined in the previous section. The time of a stride is estimated by \eqref{eqR5b}, and there is a new stride time calculated during each stride. As, by \eqref{eq2.7b}, the coordinates of the trailing leg foot tip are $(x_{f3}(t),y_{f3}(t))$, with $t\ge \gamma_k T(\phi_{02},\alpha, E_0)$, the foot tip hit the ground at the new instant of time $t=T_f$ determined by the condition
\begin{equation}
y_{f3}(t)=f(x_{f3}(t)),
\label{eq-sup4}
\end{equation}
with $t\ge \gamma_k T(\phi_{02},\alpha, E_0)$.
Therefore, you can use the model from the previous section where the duration of strides $T_f$  varies from stride to stride, depending on the shape of the surfaces.
 
\begin{figure}
\centering
\includegraphics[width=0.95\textwidth]{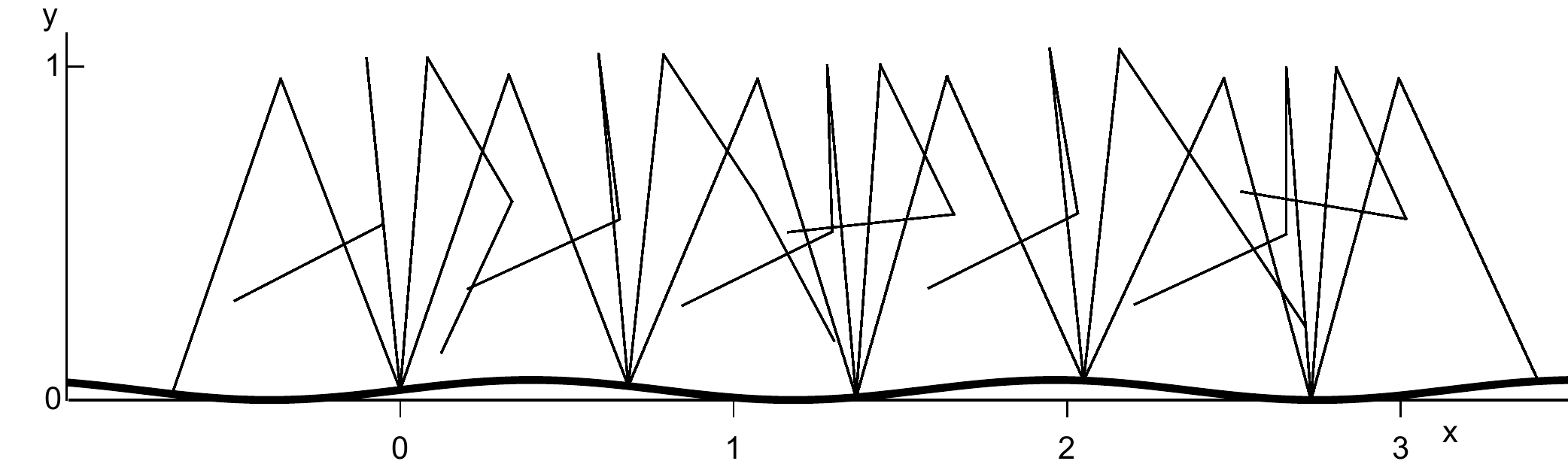}
\caption{Temporal evolution of the position of the two-legged robot on a non-planar surface described by the function \eqref{eq-sup5}, with the choice $d=0.06$ and $\omega=4$~Hz. The simulation parameters are $(x_0,y_0)=(0,f(0))$, $\phi_0=110^{\circ}$, $\alpha=70^{\circ }$, $E_0=800$~J, $g= 9.8$~ms$^{-2}$, $m=80$~kg, $\ell=1$~m, $\ell_1=0.5$~m, $\gamma_f=0.6$, $\gamma_k=0.7$ and $\alpha_r=-80^{\circ}$. With the data from figure~\ref{figR-9}, by \eqref{rough2}$, d_f=0.085\ \hbox{m}>d_{sr}=d$, and $f'(x)=d\omega/2=0.12<\tan \alpha=2.75$.
The distance between the two supporting legs is constant, $\ell_b=\ell\sqrt{2-2\cos\beta}$, regardless of the shape of the surface. }
\label{figR-11}
\end{figure}

Consider  a  surface described by the function  
\begin{equation}
{\tilde f}(x)=d(1+\sin(\omega x))/2, 
\label{eq-sup5}
\end{equation}
where the roughness parameter \eqref{rough} of this surface is  $d_{sr}=d$.
In figure~\ref{figR-11}, we show the gait movements on this non-planar surface.

\begin{figure}
\centering
\includegraphics[width=0.95\textwidth]{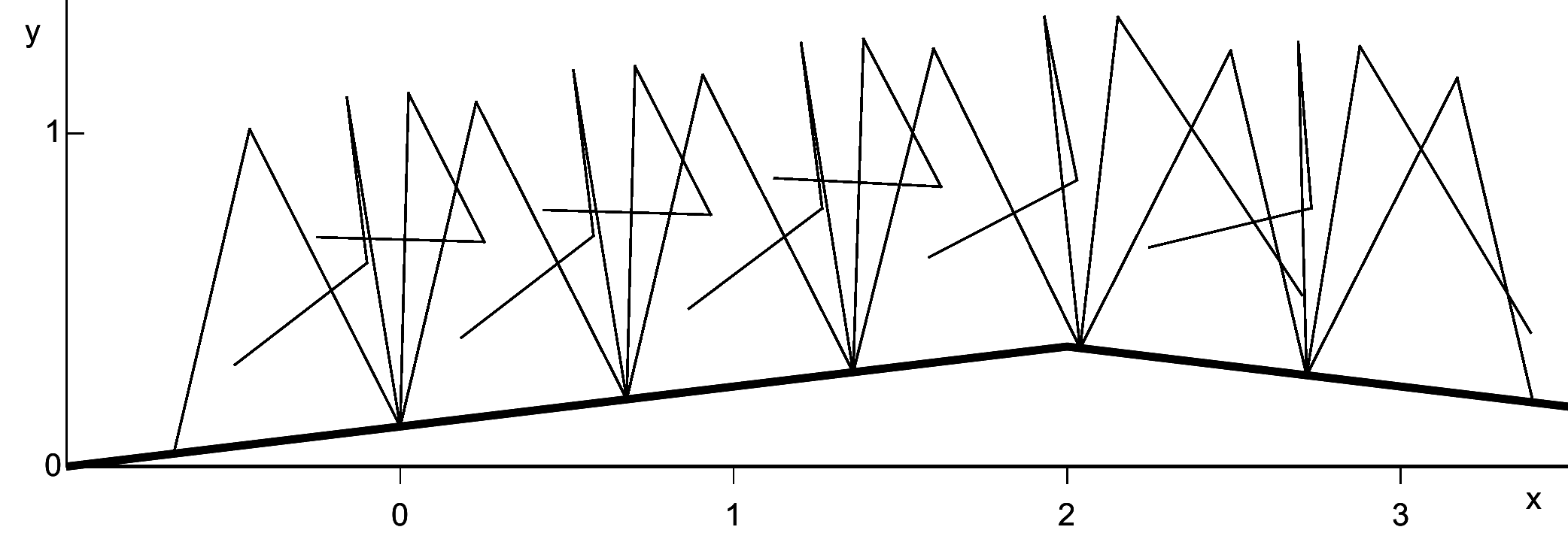}
\caption{Temporal evolution of the position of the two-legged robot on a continuous and piecewise linear surface. The shape of the surface is described by the function \eqref{eq-sup6}, with $k=0.12$. The parameters of the robotic leg are the same as in figure~\ref{figR-11}.}
\label{figR-12}
\end{figure}

In figure~\ref{figR-12}, we show the movement of the two-legged robot on inclined plane surfaces described by the function  
\begin{equation}
g(x)=\left\{\begin{array}{l}
k (x+1)\quad \hbox{if}\quad x\le 2\\
-k(x-5)\quad \hbox{if}\quad x> 2.\end{array}\right.
\label{eq-sup6}
\end{equation}

\section{Conclusions}

The solvable walking model introduced leads to the following conclusions:
\begin{itemize}
\item[a)] The articulated two-legged robot motion along a flat surface is possible provided that the initial angular speed of the standing leg obeys the condition ${\dot \phi_0}^2>2\frac{g}{\ell} (1-\cos \beta/2 )$, or, equivalently, that the initial pendulum energy obeys the condition $E_0 >mg\ell$, where $m$ is the mass of the two-legged robot, $\ell $ is the length of the legs, and $g $ is the acceleration of gravity. 
\item[b)] The robotic leg must have an internal energy supply mechanism so that, in the transition between steps, the system receives the energy $E_p$, calculated in \eqref{eqR6a} and \eqref{eqR6b}. The energy cost of walking after $n$ steps is $E_0+(n-1)E_p$. 
\item[c)] The speed of the stride of the two-legged robot is
$$
v_m=\frac{\ell \cos \alpha}{2}A\ F\left(\frac{\pi-2\alpha}{4},-4\frac{g/\ell} {A^2}\right).
$$
 where $F$ is the elliptic integral of the first kind and $A=\sqrt{2E_0/(m\ell^2)-2g/\ell}$. Under these conditions, gaiting is stable along horizontal surfaces, and the motion is reversible in time.
\item[d)] For each value of the pendular energy $E_0$ of the two-legged robot, there is a value of the opening angle $\beta$ that maximises the robot speed (figure~\ref{figR-4}). Moreover, the larger the strides, the larger the energy necessary to maintain gaiting, and the greater the energy $E_0$, the smaller the angular opening of the legs to maximise the two-legged robot speed.
\item[e)] The articulated two-legged robot can move along continuous uneven surfaces and inclined planes, provided 
that the distance from the tip of the trailing leg to the ground during the second phase of the foot movement is larger than the roughness of the ground \eqref{rough}. The distance from the tip of the trailing leg to the ground can be adjusted by choosing robot parameters (figure~\ref{figR-7}).
\end{itemize}

In conclusion, the solvable model of a two-legged robot presented here realistically describes the different types of biped motion and allows for plasticity and adaptability of gaiting. This model is stable, as minor variations on the initial conditions and robot parameters lead to neighbouring closed trajectories in phase space.

\end{document}